\documentclass[11pt]{article}
\usepackage{coling2020}
\usepackage{times}
\usepackage{url}
\usepackage{latexsym}
\usepackage{graphicx}
\usepackage{hyperref}

\colingfinalcopy 
\title{Red Dragon AI at TextGraphs 2020 Shared Task:\\
LIT : LSTM-Interleaved Transformer for Multi-Hop \\ Explanation Ranking}

\author{Yew Ken Chia \\
  Red Dragon AI  \\
  Singapore \\
  {\tt ken@reddragon.ai} \\\And
  Sam Witteveen \\
  Red Dragon AI  \\
  Singapore \\
  {\tt sam@reddragon.ai} \\\And
  Martin Andrews \\
  Red Dragon AI  \\
  Singapore \\
  {\tt martin@reddragon.ai} \\}

\date{}

\begin{document}
\maketitle
\begin{abstract}
Explainable question answering for science questions is a challenging task that requires multi-hop inference over a large set of fact sentences.
To counter the limitations of methods that view each query-document pair in isolation, we propose the LSTM-Interleaved Transformer which incorporates  cross-document interactions for improved multi-hop ranking.
The LIT architecture can leverage prior ranking positions in the re-ranking setting.
Our model is competitive on the current leaderboard for the TextGraphs 2020 shared task, achieving a test-set MAP of 0.5607, and would have gained third place had we submitted before the competition deadline.
Our code implementation is made available at
\url{https://github.com/mdda/worldtree_corpus/tree/textgraphs_2020}
\end{abstract}

\section{Introduction}
\label{intro}

\blfootnote{
    %
    %
    %
    %
    %
    %
    \hspace{-0.65cm}  
    This work is licensed under a Creative Commons 
    Attribution 4.0 International License.
    License details:
    \url{http://creativecommons.org/licenses/by/4.0/}.
}


\begin{figure}[h]
\centering
    \begin{minipage}{.5\textwidth}
        \centering
        \includegraphics[width=0.8\textwidth]{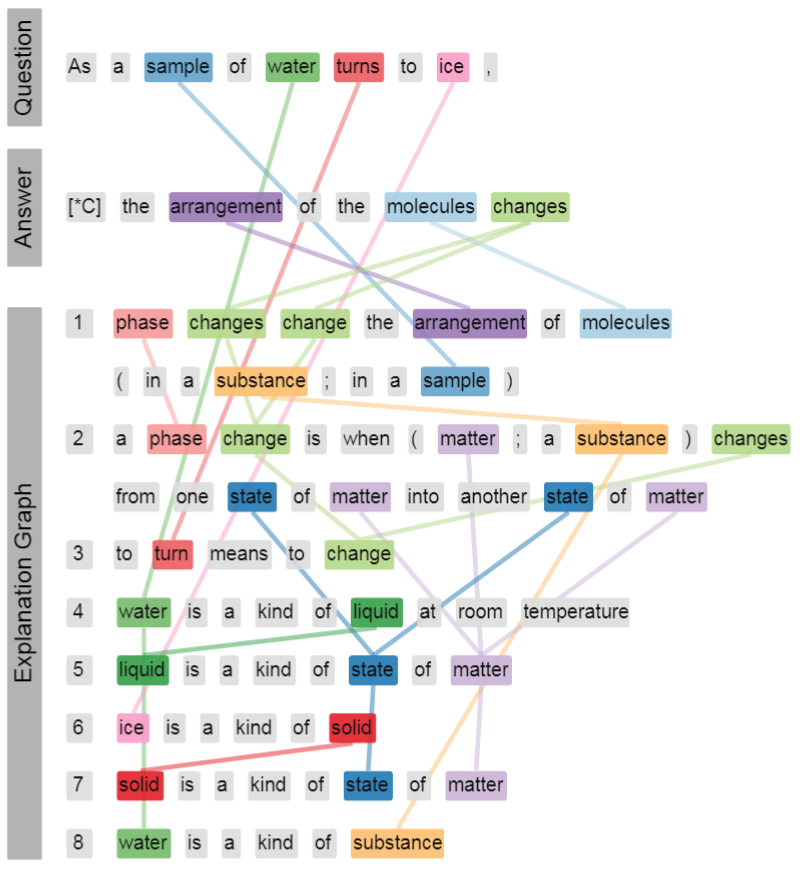}
        \caption{Example of a single question and ground-truth explanation facts in WorldTree V2 dataset.}
        \label{fig:graph_example}
    \end{minipage}%
    \begin{minipage}{.5\textwidth}
        \centering
        \includegraphics[width=1.0\textwidth]{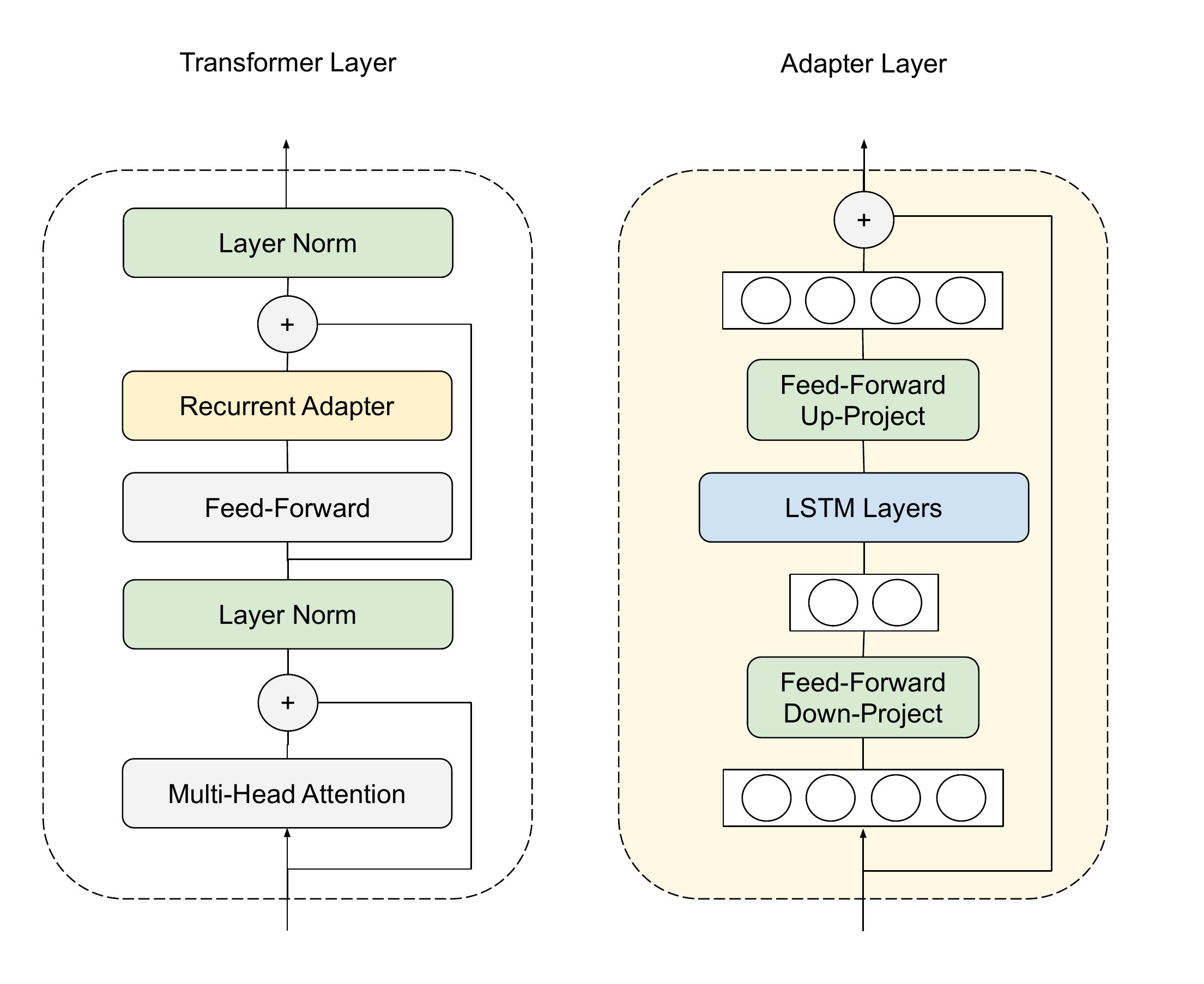}
        \caption{Proposed LIT architecture}
        \label{fig:architecture}
    \end{minipage}
\end{figure}


Complex question answering often requires reasoning over many evidence documents, which is known as multi-hop inference.
Existing datasets such as Wikihop \cite{welbl-etal-2018-constructing}, OpenBookQA \cite{OpenBookQA2018}, QASC \cite{khot2020qasc}, are limited due to artificial questions and short aggregation, requiring less than 3 facts.
In comparison, TextGraphs \cite{textgraphs} uses WorldTree V2 \cite{xie-etal-2020-worldtree} which is the largest available dataset that requires combining an average of 6 and up to 16 facts in order to generate an explanation for complex science questions.
The dataset contains 5k questions that require knowledge in core science as well as common sense.
Figure \ref{fig:graph_example} shows an example question from the WorldTree dataset.
The evaluation for this dataset is framed as a ranking objective over a large set of 9k science facts, and models are scored based on the MAP metric over the predicted rank ordering.
Multi-hop inference encounters significant noise or ``distraction'' documents in the process and this challenge is known as semantic drift \cite{fried-etal-2015-higher}.
Compared to WorldTree V1 \cite{JANSEN18.81}, WorldTree V2 has more examples but is more challenging as the larger pool of science facts presents a greater risk of semantic drift.
\\
\\
Neural information retrieval models such as DPR \cite{karpukhin2020dense}, RAG \cite{lewis2020retrieval}, and ColBERT \cite{khattab2020colbert} that assume query-document independence use a language model to generate sentence representations for the query and document separately.
The advantage of this late-interaction approach is efficient inference as the sentence representations can be computed beforehand and optimized lookup methods such as FAISS \cite{JDH17} exist for this purpose.
However, the late-interaction compromises on deeper semantic understanding possible with language models.
Early-interaction approaches such as TFR-BERT \cite{han2020learning} instead concatenate the query and document before generating a unified sentence representation. 
This approach is more computationally expensive but is attractive for re-ranking over a limited number of documents.
However, the previous approaches consider each query-document pair in isolation. 
This forgoes any cross-document interaction which can leverage additional knowledge sources or benefit the ranking objective. 
Other work \cite{pasumarthi2019selfattentive,pobrotyn2020context,sun2020modeling} facilitate cross-document interactions through self-attention mechanisms.
However, the cross-document interaction is only applied after the feature extraction step and cannot leverage the language understanding potential in earlier language model layers.
\\
\\
The most straightforward loss for the document ranking objective is Binary Crossentropy where each document is ranked according to the binary classification probability of being within the gold explanation set.  However, there have been recent progress in differentiable losses to optimize directly for the ranking objective \cite{wang2018lambdaloss,revaud2019learning,engilberge2019sodeep}.
In this work, we also compare the benefits of each loss for multi-hop ranking.
\\
\\
The main contributions of this work are:
\begin{enumerate}
    \item We show that conventional information retrieval-based methods are still a strong baseline and propose I-BM25, an iterative retrieval method that improves inference speed and recall by emulating multi-hop retrieval.
    \item We propose a hierarchical LSTM-interleaved transformer (LIT) architecture that maximizes early cross-document interactions for improved multi-hop re-ranking.
    \item We provide empirical comparisons of training with different loss functions and show that Binary Crossentropy loss is simple yet may outperform differentiable ranking losses.
\end{enumerate}

\section{Models}
\begin{figure}[t]
    \centering
    \includegraphics[width=1.0\textwidth]{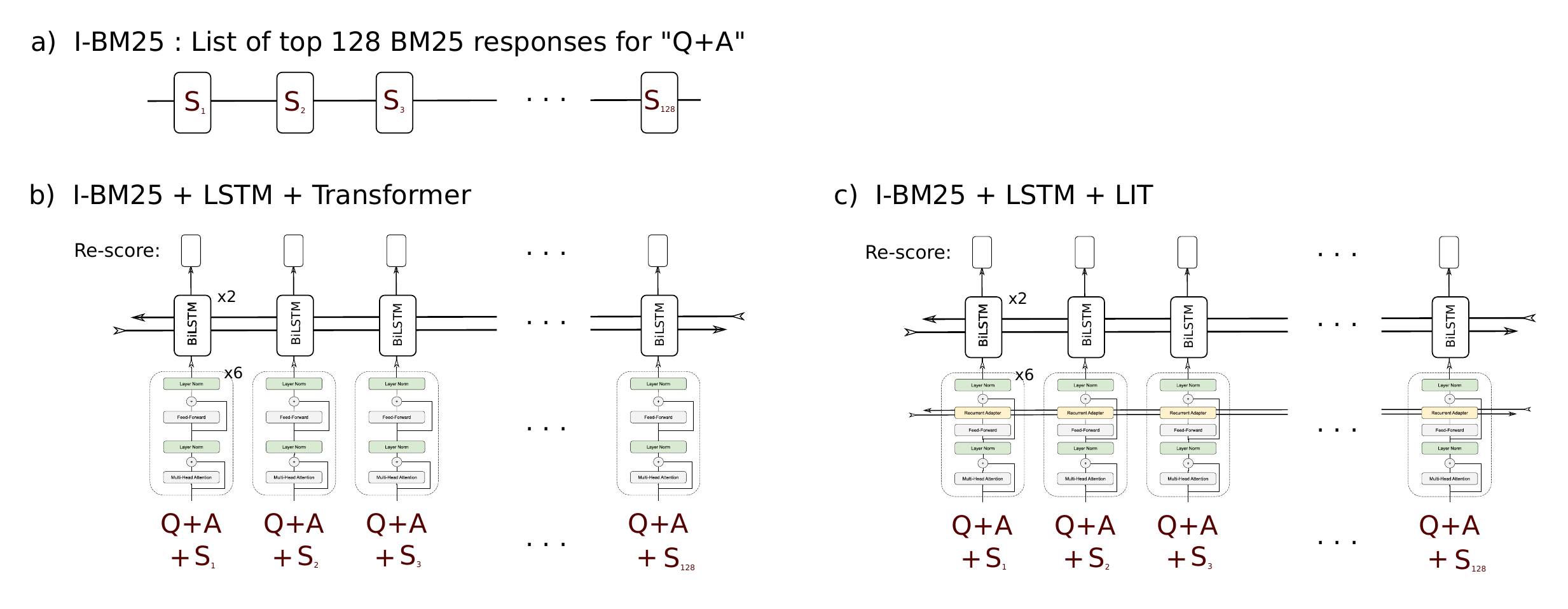}
    \caption{Overview of 3 architectures}
    \label{fig:systems}
\end{figure}

Three different system architectures are described here, with overall schemes illustrated in Figure \ref{fig:systems} for comparison.

\subsection{Iterative BM25 Retrieval}
Chia et al \shortcite{chia-etal-2019-red} showed that conventional information retrieval methods can be a strong baseline when modified to suit the multi-hop inference objective.
However, this method is limited due to computationally expensive inference and sensitivity to noise and semantic drift.
We propose an iterative retrieval method `I-BM25' that performs inference in a fraction of the time and reduces semantic drift, resulting in a even stronger baseline retrieval method.
For preprocessing, we use spaCy \cite{spacy2} for tokenization, lemmatization and stopword removal.
Compared to Chia et al \shortcite{chia-etal-2019-red} which processes each new candidate one at a time, I-BM25 processes $2^n$ candidates in the $n$-th iteration.
The algorithm is as follows:

\begin{enumerate}
    \item Sparse document vectors are pre-computed for all questions and explanation candidates.
    \item For each question, the closest $n$ explanation candidates by cosine proximity are selected and their vectors are aggregated by a $max$ operation.
    The aggregated vector is down-scaled and used to update the query vector through a $max$ operation.
    \item The previous step is repeated for increasing values of $n$ until there are no candidate explanations remaining.
\end{enumerate}

\subsection{LSTM-After Transformer for Re-Ranking}
BERT is a pre-trained language model that is widely adapted and fine-tuned for many downstream NLP tasks. 
Due to computational constraints, we use DistilBERT \cite{sanh2020distilbert} which has 40\% fewer parameters and comparable performance.
In sequence-level tasks such as text classification, a [CLS] token is a special token inserted at the front of the sequence.
The latent representation of the token is passed to a feed-forward network for prediction.
We append an LSTM \cite{HochSchm97} module with 2 layers that operate on the [CLS] vectors of the last layer of BERT (similar in principle to McCann et al \shortcite{mccann2018natural}).
This hierarchical structure allows the transformer to perform cross-document reasoning and knowledge reference.
The LSTM layers enable the model to be rank-aware when used in the re-ranking setting.
For re-ranking, the top 128 predictions from I-BM25 are passed to the LSTM-After Transformer which performs binary classification for each document.

\subsection{LSTM-Interleaved Transformer for Re-Ranking}

TextGraphs is a challenging task which requires complex multi-hop reasoning, but information retrieval methods are surprisingly strong baselines. To enhance cross-document interaction and leverage language representations in earlier transformer layers, we interleave adapters \cite{houlsby2019parameter} into the architecture which are recurrent instead of merely feed-forward.
The LSTM-adapter modules in Figure \ref{fig:architecture} operate on the latent representation at the [CLS] position of each document \textit{at each layer} of the transformer.
After each transformer layer, the [CLS] latent representations for each input document are first down-projected, passed to the LSTM layers and finally up-projected and fed into the next transformer layer. 
Compared to \cite{houlsby2019parameter}, the LIT architecture is fully trainable and makes the transformer architecture more expressive by enabling cross-document reasoning which was previously not possible.
Apart from LSTM, we also tested GCN \cite{gcn} and Self-Attention \cite{selfattention} layers but had limited success in achieving competitive performance from them.

\section{Experiments}

\begin{table}[h!]
\centering
\begin{tabular}{|l  | r r |} 
 \hline
 Model & Dev MAP & Test MAP \\
 \hline
 BM25 & 0.4615 & \\
 Iterative BM25 \cite{chia-etal-2019-red} & 0.4704 & \\
 I-BM25 & 0.4861 & 0.4745 \\ 
 I-BM25 + LSTM + Transformer & 0.5470 & 0.5294 \\
 I-BM25 + LIT & 0.5680 & 0.5607 \\
\hline
\end{tabular}
\caption{Main score comparison on WorldTree V2 dataset}
\label{table:1}
\end{table}

Table \ref{table:1} shows that I-BM25 is a strong information retrieval method that can be a drop-in replacement for previous information retrieval methods.
The results also show the advantage of the LIT architecture in interleaving LSTM layers between transformer layers, rather than after the last transformer layer.

\begin{table}[h!]
\centering
\begin{tabular}{|l  | r |} 
 \hline
 Loss Function & Dev MAP \\
 \hline
 LambdaLoss & 0.4970 \\
 APLoss & 0.5187 \\
 Binary Crossentropy & 0.5680 \\
\hline
\end{tabular}
\caption{Loss function comparison on WorldTree V2 dataset}
\label{table:2}
\end{table}

The results of optimization using 3 different loss objectives are shown in Table \ref{table:2}.
Surprisingly, the direct ranking-loss oriented objectives were less effective in reducing the final evaluation MAP score, which is potentially due to the bucketisation approximation used in the APLoss calculations not being appropriately pre-scaled in our experiments.
In this case, the training may require different hyper-parameters to converge optimally.
Another potential explanation is that these ranking losses may be sub-optimal (when used as a training objective) when many documents have very similar underlying scores which is the case here.

\subsection{Notes}
Further to our experience last year, we included preprocessing steps to isolate the branching `combo' statements (which essentially contain OR clauses between different noun phrases, for instance).  This step remains in our codebase, but we did not exploit it fully, since a full treatment would require the isolation of which `combo branch' is taken by each gold statement in the training set.

\section{Discussion}
Other architectures that we explored included Graph neural network (GNN) methods, however we had insufficient time to tune these for the multi-hop explanation task herein.  Surprisingly, our simple LSTM methods (which can be viewed as a linear graph that performs message-passing along the list of results ordered by the I-BM25 method) already provided a competitive method.  We estimate that next year's competition will {\textit require} the use of graph-based methods, due to their greater expressive power.  

\section{Conclusion}
The LIT architecture is a simple yet powerful adaptation of the Transformer architecture to learn better cross-document interactions for multi-hop ranking.
The structure can be easily integrated with any transformer language model to enable cross-referencing of knowledge statements and improved ranking performance.
For example, LIT can be a drop-in encoder for other multi-hop question answering datasets such as HotPotQA \cite{hotpotqa}.
When applied to the challenging WorldTree V2 dataset, LIT achieves competitive performance with current state-of-the-art models despite a smaller footprint.
We envision that this architecture can be beneficial to many NLP tasks which require multi-hop reasoning over documents.

\bibliographystyle{coling}
\bibliography{coling2020}

\begin{thebibliography}{}

\bibitem[\protect\citename{Chia \bgroup et al.\egroup
  }2019]{chia-etal-2019-red}
Yew~Ken Chia, Sam Witteveen, and Martin Andrews.
\newblock 2019.
\newblock Red dragon {AI} at {T}ext{G}raphs 2019 shared task: Language model
  assisted explanation generation.
\newblock In {\em Proceedings of the Thirteenth Workshop on Graph-Based Methods
  for Natural Language Processing (TextGraphs-13)}, pages 85--89, Hong Kong,
  November. Association for Computational Linguistics.

\bibitem[\protect\citename{Engilberge \bgroup et al.\egroup
  }2019]{engilberge2019sodeep}
Martin Engilberge, Louis Chevallier, Patrick P{\'e}rez, and Matthieu Cord.
\newblock 2019.
\newblock Sodeep: a sorting deep net to learn ranking loss surrogates.
\newblock In {\em Proceedings of the IEEE Conference on Computer Vision and
  Pattern Recognition}, pages 10792--10801.

\bibitem[\protect\citename{Fried \bgroup et al.\egroup
  }2015]{fried-etal-2015-higher}
Daniel Fried, Peter Jansen, Gustave Hahn-Powell, Mihai Surdeanu, and Peter
  Clark.
\newblock 2015.
\newblock Higher-order lexical semantic models for non-factoid answer
  reranking.
\newblock {\em Transactions of the Association for Computational Linguistics},
  3:197--210.

\bibitem[\protect\citename{Han \bgroup et al.\egroup }2020]{han2020learning}
Shuguang Han, Xuanhui Wang, Mike Bendersky, and Marc Najork.
\newblock 2020.
\newblock Learning-to-rank with bert in tf-ranking.
\newblock {\em arXiv preprint arXiv:2004.08476}.

\bibitem[\protect\citename{Hochreiter and Schmidhuber}1997]{HochSchm97}
Sepp Hochreiter and Jürgen Schmidhuber.
\newblock 1997.
\newblock Long short-term memory.
\newblock {\em Neural Computation}, 9(8):1735--1780.

\bibitem[\protect\citename{Honnibal and Montani}2017]{spacy2}
Matthew Honnibal and Ines Montani.
\newblock 2017.
\newblock {spaCy 2}: Natural language understanding with {B}loom embeddings,
  convolutional neural networks and incremental parsing.
\newblock To appear.

\bibitem[\protect\citename{Houlsby \bgroup et al.\egroup
  }2019]{houlsby2019parameter}
Neil Houlsby, Andrei Giurgiu, Stanislaw Jastrzebski, Bruna Morrone, Quentin
  de~Laroussilhe, Andrea Gesmundo, Mona Attariyan, and Sylvain Gelly.
\newblock 2019.
\newblock Parameter-efficient transfer learning for nlp.
\newblock In {\em ICML}.

\bibitem[\protect\citename{Jansen and Ustalov}2020]{textgraphs}
Peter Jansen and Dmitry Ustalov.
\newblock 2020.
\newblock {TextGraphs~2020 Shared Task on Multi-Hop Inference for Explanation
  Regeneration}.
\newblock In {\em Proceedings of the Graph-based Methods for Natural Language
  Processing (TextGraphs)}. Association for Computational Linguistics.

\bibitem[\protect\citename{Jansen \bgroup et al.\egroup }2018]{JANSEN18.81}
Peter Jansen, Elizabeth Wainwright, Steven Marmorstein, and Clayton Morrison.
\newblock 2018.
\newblock {WorldTree: A Corpus of Explanation Graphs for Elementary Science
  Questions supporting Multi-hop Inference}.
\newblock In Nicoletta Calzolari~(Conference chair), Khalid Choukri,
  Christopher Cieri, Thierry Declerck, Sara Goggi, Koiti Hasida, Hitoshi
  Isahara, Bente Maegaard, Joseph Mariani, Hélène Mazo, Asuncion Moreno, Jan
  Odijk, Stelios Piperidis, and Takenobu Tokunaga, editors, {\em Proceedings of
  the Eleventh International Conference on Language Resources and Evaluation
  (LREC 2018)}, Miyazaki, Japan, May 7-12, 2018. European Language Resources
  Association (ELRA).

\bibitem[\protect\citename{Johnson \bgroup et al.\egroup }2017]{JDH17}
Jeff Johnson, Matthijs Douze, and Herv{\'e} J{\'e}gou.
\newblock 2017.
\newblock Billion-scale similarity search with gpus.
\newblock {\em arXiv preprint arXiv:1702.08734}.

\bibitem[\protect\citename{Karpukhin \bgroup et al.\egroup
  }2020]{karpukhin2020dense}
Vladimir Karpukhin, Barlas O{\u{g}}uz, Sewon Min, Ledell Wu, Sergey Edunov,
  Danqi Chen, and Wen-tau Yih.
\newblock 2020.
\newblock Dense passage retrieval for open-domain question answering.
\newblock {\em arXiv preprint arXiv:2004.04906}.

\bibitem[\protect\citename{Khattab and Zaharia}2020]{khattab2020colbert}
Omar Khattab and Matei Zaharia.
\newblock 2020.
\newblock Colbert: Efficient and effective passage search via contextualized
  late interaction over bert.
\newblock {\em arXiv preprint arXiv:2004.12832}.

\bibitem[\protect\citename{Khot \bgroup et al.\egroup }2020]{khot2020qasc}
Tushar Khot, Peter Clark, Michal Guerquin, Peter Jansen, and Ashish Sabharwal.
\newblock 2020.
\newblock Qasc: A dataset for question answering via sentence composition.
\newblock In {\em AAAI}, pages 8082--8090.

\bibitem[\protect\citename{Kipf and Welling}2017]{gcn}
Thomas Kipf and M.~Welling.
\newblock 2017.
\newblock Semi-supervised classification with graph convolutional networks.
\newblock {\em ArXiv}, abs/1609.02907.

\bibitem[\protect\citename{Lewis \bgroup et al.\egroup
  }2020]{lewis2020retrieval}
Patrick Lewis, Ethan Perez, Aleksandara Piktus, Fabio Petroni, Vladimir
  Karpukhin, Naman Goyal, Heinrich K{\"u}ttler, Mike Lewis, Wen-tau Yih, Tim
  Rockt{\"a}schel, et~al.
\newblock 2020.
\newblock Retrieval-augmented generation for knowledge-intensive nlp tasks.
\newblock {\em arXiv preprint arXiv:2005.11401}.

\bibitem[\protect\citename{McCann \bgroup et al.\egroup
  }2018]{mccann2018natural}
Bryan McCann, Nitish~Shirish Keskar, Caiming Xiong, and Richard Socher.
\newblock 2018.
\newblock The natural language decathlon: Multitask learning as question
  answering.

\bibitem[\protect\citename{Mihaylov \bgroup et al.\egroup
  }2018]{OpenBookQA2018}
Todor Mihaylov, Peter Clark, Tushar Khot, and Ashish Sabharwal.
\newblock 2018.
\newblock Can a suit of armor conduct electricity? a new dataset for open book
  question answering.
\newblock In {\em EMNLP}.

\bibitem[\protect\citename{Parikh \bgroup et al.\egroup }2016]{selfattention}
Ankur~P. Parikh, Oscar T{\"a}ckstr{\"o}m, Dipanjan Das, and Jakob Uszkoreit.
\newblock 2016.
\newblock A decomposable attention model for natural language inference.
\newblock {\em ArXiv}, abs/1606.01933.

\bibitem[\protect\citename{Pasumarthi \bgroup et al.\egroup
  }2019]{pasumarthi2019selfattentive}
Rama~Kumar Pasumarthi, Xuanhui Wang, Michael Bendersky, and Marc Najork.
\newblock 2019.
\newblock Self-attentive document interaction networks for permutation
  equivariant ranking.

\bibitem[\protect\citename{Pobrotyn \bgroup et al.\egroup
  }2020]{pobrotyn2020context}
Przemys{\l}aw Pobrotyn, Tomasz Bartczak, Miko{\l}aj Synowiec, Rados{\l}aw
  Bia{\l}obrzeski, and Jaros{\l}aw Bojar.
\newblock 2020.
\newblock Context-aware learning to rank with self-attention.
\newblock {\em arXiv preprint arXiv:2005.10084}.

\bibitem[\protect\citename{Revaud \bgroup et al.\egroup
  }2019]{revaud2019learning}
Jerome Revaud, Jon Almaz{\'a}n, Rafael~S Rezende, and Cesar Roberto~de Souza.
\newblock 2019.
\newblock Learning with average precision: Training image retrieval with a
  listwise loss.
\newblock In {\em Proceedings of the IEEE International Conference on Computer
  Vision}, pages 5107--5116.

\bibitem[\protect\citename{Sanh \bgroup et al.\egroup
  }2020]{sanh2020distilbert}
Victor Sanh, Lysandre Debut, Julien Chaumond, and Thomas Wolf.
\newblock 2020.
\newblock Distilbert, a distilled version of bert: smaller, faster, cheaper and
  lighter.

\bibitem[\protect\citename{Sun and Duh}2020]{sun2020modeling}
Shuo Sun and Kevin Duh.
\newblock 2020.
\newblock Modeling document interactions for learning to rank with regularized
  self-attention.
\newblock {\em arXiv preprint arXiv:2005.03932}.

\bibitem[\protect\citename{Wang \bgroup et al.\egroup
  }2018]{wang2018lambdaloss}
Xuanhui Wang, Cheng Li, Nadav Golbandi, Michael Bendersky, and Marc Najork.
\newblock 2018.
\newblock The lambdaloss framework for ranking metric optimization.
\newblock In {\em Proceedings of the 27th ACM International Conference on
  Information and Knowledge Management}, pages 1313--1322.

\bibitem[\protect\citename{Welbl \bgroup et al.\egroup
  }2018]{welbl-etal-2018-constructing}
Johannes Welbl, Pontus Stenetorp, and Sebastian Riedel.
\newblock 2018.
\newblock Constructing datasets for multi-hop reading comprehension across
  documents.
\newblock {\em Transactions of the Association for Computational Linguistics},
  6:287--302.

\bibitem[\protect\citename{Xie \bgroup et al.\egroup
  }2020]{xie-etal-2020-worldtree}
Zhengnan Xie, Sebastian Thiem, Jaycie Martin, Elizabeth Wainwright, Steven
  Marmorstein, and Peter Jansen.
\newblock 2020.
\newblock {W}orld{T}ree v2: A corpus of science-domain structured explanations
  and inference patterns supporting multi-hop inference.
\newblock In {\em Proceedings of the 12th Language Resources and Evaluation
  Conference}, pages 5456--5473, Marseille, France, May. European Language
  Resources Association.

\bibitem[\protect\citename{Yang \bgroup et al.\egroup }2018]{hotpotqa}
Z.~Yang, Peng Qi, Saizheng Zhang, Yoshua Bengio, William~W. Cohen,
  R.~Salakhutdinov, and Christopher~D. Manning.
\newblock 2018.
\newblock Hotpotqa: A dataset for diverse, explainable multi-hop question
  answering.
\newblock {\em ArXiv}, abs/1809.09600.

\end{thebibliography}


\end{document}